# Local Frequency Domain Transformer Networks for Video Prediction

Hafez Farazi, Jan Nogga, Sven Behnke

*University of Bonn, Computer Science Institute VI, Autonomous Intelligent Systems*

Friedrich-Hirzebruch-Allee 5, 53115 Bonn, Germany

{farazi, nogga, behnke}@ais.uni-bonn.de

*Abstract*—Video prediction is commonly referred to as forecasting future frames of a video sequence provided several past frames thereof. It remains a challenging domain as visual scenes evolve according to complex underlying dynamics, such as the camera's egocentric motion or the distinct motility per individual object viewed. These are mostly hidden from the observer and manifest as often highly non-linear transformations between consecutive video frames. Therefore, video prediction is of interest not only in anticipating visual changes in the real world but has, above all, emerged as an unsupervised learning rule targeting the formation and dynamics of the observed environment. Many of the deep learning-based state-of-the-art models for video prediction utilize some form of recurrent layers like Long Short-Term Memory (LSTMs) or Gated Recurrent Units (GRUs) at the core of their models. Although these models can predict the future frames, they rely entirely on these recurrent structures to simultaneously perform three distinct tasks: extracting transformations, projecting them into the future, and transforming the current frame. In order to completely interpret the formed internal representations, it is crucial to disentangle these tasks. This paper proposes a fully differentiable building block that can perform all of those tasks separately while maintaining interpretability. We derive the relevant theoretical foundations and showcase results on synthetic as well as real data. We demonstrate that our method is readily extended to perform motion segmentation and account for the scene's composition, and learns to produce reliable predictions in an entirely interpretable manner by only observing unlabeled video data.

## I. Introduction

A powerful video predictor presupposes a capability to model both scene composition and dynamics. Many recent video prediction approaches resort to deep architectures that require a vast number of parameters to extract relevant features from observed sequences, which raise many scalability issues. Large networks take days to train on even synthetic datasets, which renders exploration of new ideas more difficult than lightweight differentiable models. Lightweight models are trained relatively swiftly and need fewer training data points. More importantly, they do not tend to overfit the training set, therefore generalizing to novel data better than their heavy counterparts.

This work was funded by grant BE 2556/16-1 of the German Research Foundation (DFG).

Additionally, there is an inherent lack of interpretability to complex models. Even once enough data is observed, and predictions are visually plausible, there is no straightforward way to understand the encodings in hidden states of such models, be it for retracing their decisions or explicitly retrieving information on objects or people interacting in the video data, which is often the ultimate target. Note that video prediction is often used as a proxy task in a self-supervised context to form useful and meaningful representations for another downstream task.

One way to address these issues is to pre-structure the models based on domain knowledge. Of course, manually engineering every aspect of video prediction is not possible, and one has to find the proper balance between nature—inductive bias, which is optimized on an evolutionary time scale—and nurture, learning from own experience.

Many of the deep learning-based video prediction models utilize recurrent layers like or GRUs at their models' core. These are essentially black boxes, performing three different tasks simultaneously and in an interwoven fashion - extracting transformations, projecting them into the future, and transforming the visual content. We argue that to understand the model's internal representation, it is crucial to untangle these tasks. Our proposed model performs all three tasks sequentially and separately.

We introduce a flexible and content-independent transformer model that can perform local predictions at selectable degrees of sparsity by effectively applying the Frequency Domain Transformer Network architecture locally [1]. In this way, our model relies on few trainable parameters and is fully interpretable. The corresponding prediction pipeline is end-to-end trainable, and learning, where applied, is entirely explainable. It is also lightweight and flexible, enabling its use as a building block at the core of sophisticated video prediction systems. We demonstrate this by including such a module in a system that can account for depth in an observed scene, achieving competitive results on synthetic and real datasets while also learning motion segmentation in a self-supervised manner.

The code and dataset of this paper is publicly available on github [*].

## II. Related Work

While many diverse approaches to video prediction have been suggested, the most effective ones apply deep learning to form abstract representations of scene contents and observed transformations. A well-known successful example is Video Ladder Network (VLN) [2], an extension of Ladder Networks [3] that occupies a recurrent lateral connection at each level, modeling transformations at that level of abstraction, the bottom-most representing the video frames. Reciprocally, PredRNN++ [4] comprises a stack of LSTM modules, the output of each directed into the subsequent one, with a frame prediction formed at the top. Advancing neural plausibility, PredNet [5] implements a hierarchical architecture that learns a generative model of the input per layer. Only the discrepancy from the expected input is passed upwards, and thereby concepts from predictive coding are actualized. In an extension of this idea, HPNet [6] also draws from associative coding, adding a direct upward stream of spatiotemporal feature encodings extracted by 3D-convolutions. Here, the feedback path is directed to an LSTM at each level. Beyond producing plausible future frames, both aforementioned ideas highlight the exciting potential of video prediction tasks in investigating models of cortical processing.

In stark contrast, other approaches are by design completely nescient regarding image contents. PGP [7], [8] integrates a gated auto-encoder and the transformation model of RAE [9] to learn encodings of global linear image transformations between consecutive frames. Frequency Domain Transformer Networks (FDTN) [1], on the other hand, recover shifts between consecutive frames from the corresponding Fourier partners' phase difference. Finally, van Amersfoort et al. [10] represent the image transformation as a set of affine transformations locally estimated in a sliding window manner and smoothed by a series of convolutional layers. In each of these techniques, the estimated image transformation can be globally or locally applied to produce future frames. The local transform model has a larger capacity to represent different image transformations, but the global models are readily extended to represent higher-order derivatives as transformations of transformations.

In the scope of the coarse taxonomy of video prediction techniques implied above, the idea presented in this work is appendant to the latter class. However, we are convinced that appreciation of scene composition is no less vital than exploiting redundancies in the image transformations. We hope to bridge this gap without losing interpretability by ensuring that and showing how our model can be combined with other functional components.

On another note, the representation of image transformations developed in the following is visualized by sparse vector fields. While similar, we warn against construing these as equivalent to optical flow. Even though comparable representations could be calculated from the results of classical optical flow algorithms or the output of deep models like FlowNet 2.0 [11], neither is suited for use in our pipeline, which has to remain slim, differentiable, and explainable.

## III. Local Frequency Domain Transformer Networks (LFDTN)

The previous results of FDTN-based approaches [1], [12] are encouraging and illustrate the benefits of an explicit transformer model at the core of an architecture learning to comprehend the underlying video scene. It allows light-weight models that can reason on the probable composition of the observed environment. However, the transformer model using a global phase-add prediction is ultimately confined to describing the observed changes with one translation per identified layer.

To address this limitation, we re-examine the transformation model. Using a local method similar to the procedure by Amersfoort et al. [10], we describe the transformation between video frames as local translational image movement measured by the phase differences between local cells in the observed images.

This remains differentiable, allowing its use at the base of models that can comprehend scene parameters. Unlike the affine transformations estimated in the spatial domain used in [10], this approach can be extended to represent higher-order derivatives similar to the technique of Michalski et al. [13], via differences of differences as described by Farazi [1] for the global case and interpreted by visualizing the local shifts.

We can move to apply our analysis to overlapping cells covering an image on a regular grid by first formalizing these local cells' extraction. Consider a $U \times V$ image $x$ covered by overlapping cells of size $N \times N$. We refer to $N$ as the *window size*. With a *stride* of $H$, the overlap between the cells along the coordinate axes is $N - H$, and restricting $H \leq N$ guarantees that each pixel in $x$ is represented in at least one cell. Additionally, we impose the constraint that $H$ must be a divisor of $U-1$ and $V-1$. This allows placement of the grid of cells such that the first cell is anchored on the top-left pixel, while the final cell is centered on the right-most pixel in the bottom row of the image, ensuring symmetrical coverage. Consequently, the image requires padding with at least $\lfloor \frac{N}{2} \rfloor$ pixels to accommodate cells which reach beyond the image border. In this scenario, image pixels near the border are viewed less often than those near the center; thus, we extend the grid by $k$ further cells, with:

$$k = \arg\max_{i}(\{i \mid iH \leq \lfloor \frac{N}{2} \rfloor\}). \tag{1}$$

This ensures consistent coverage of image pixels, but requires an additional $kH$ pixels of image padding. In this manner, $L_U := \frac{U-1}{H} + 1 + 2k$ and $L_V := \frac{V-1}{H} + 1 + 2k$ cells are extracted along the $y$ and $x$ axis, respectively. The total number of extracted cells is referred to as $L := L_U L_V$.

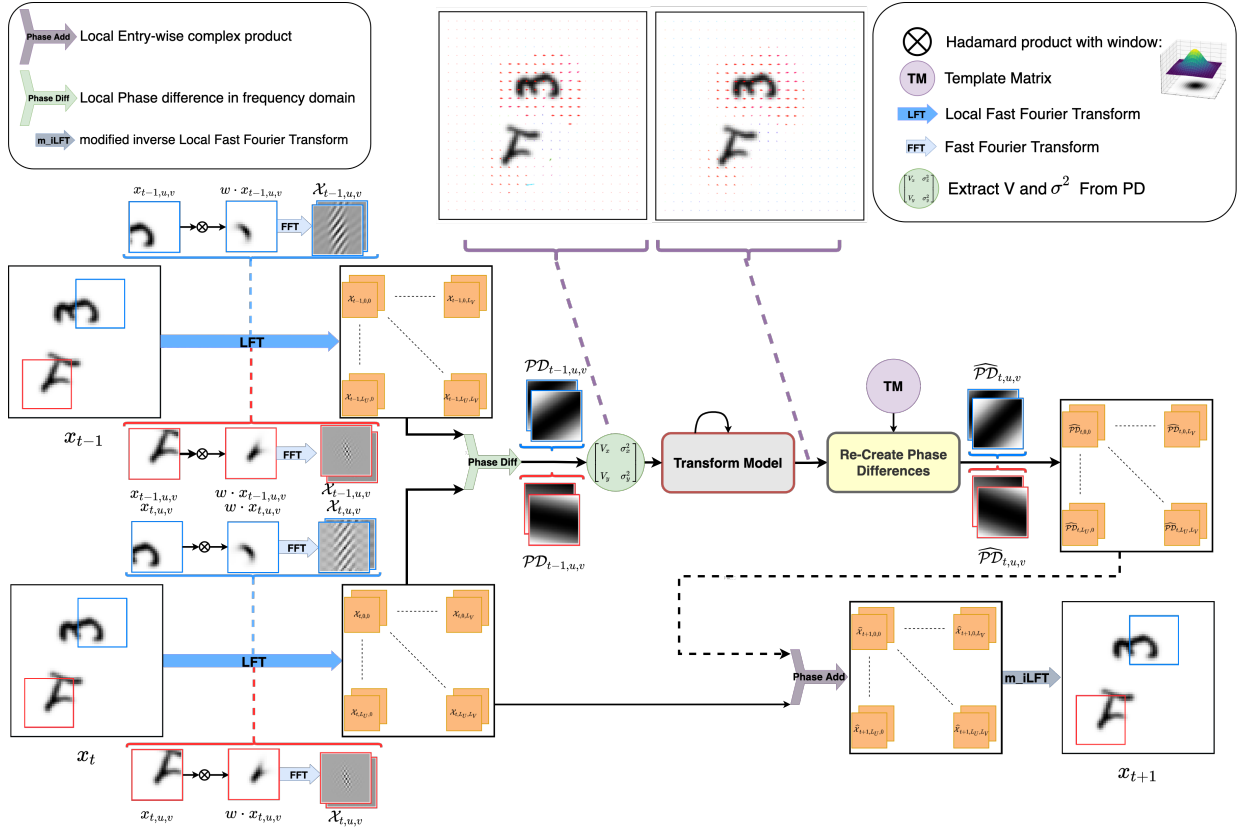

Fig. 1: Our proposed prediction scheme. Overlapping cells covering the image are extracted, tapered, and Fourier-transformed for consecutive frames. The local phase differences between the results are calculated, then interpreted as local shifts, then are spatiotemporally adjusted by the "transform model". The output is converted back to a phase-based description, phase-added to the local Fourier domain representation of the most recent frame, and finally used to construct a prediction of the next one by inverse Fourier-transform and a consecutive overlap-add procedure. The two extracted local cells from the input, highlighted in red and blue, exemplify a local view of the process.

We describe the cell at $(u,v)$, referred to here as $x_{u,v}$, by defining a window $w$ shaped $N \times N$, shifting it to the correct location, and then multiplying with $x$ such that:

$$x_{u,v}[n,m] = x[n,m] \cdot w[n-u \cdot H, m-v \cdot H] \quad (2)$$

For an adequate notation for cell extraction, it is sufficient to suppose that $w$ is a rectangle window, but note that we need to revisit this term shortly!

Following the cell extraction step, the corresponding local FFTs are given by:

$$C = e^{-j\frac{2\pi}{N}(\omega_1(n-uH)+\omega_2(m-vH))} \quad (3)$$

$$\mathcal{X}_{u,v}[\omega_1,\omega_2] = \sum_{n=uH}^{uH+N-1}\sum_{m=vH}^{vH+N-1} x_{u,v}[n,m] C \quad (4)$$

$$= \sum_{n=uH}^{uH+N-1}\sum_{m=vH}^{vH+N-1} x[n,m]w[n-uH,m-vH]C \quad (5)$$

$$= \sum_{n=0}^{N-1}\sum_{m=0}^{N-1} x[n+uH,m+vH]w[n,m]e^{-j\frac{2\pi}{N}(\omega_1 n+\omega_2 m)} \quad (6)$$

In tandem with the previous step, this defines a two-dimensional version of the *Short-Time Fourier Transform* (STFT) [14]. Since it is a self-evident extension of the STFT, renaming would not be warranted. Still, individual images' indices are spatial rather than temporal, so this is not a fitting denomination. In the following, we refer to this process as a *Local Fourier Transform* (LFT), summarized in algorithm 1. This type of analysis has already been successfully applied by Lazar et al. [15] to detect motion like a Reichardt detector.

Given $\mathcal{X}_{t-1,u,v}$ and $\mathcal{X}_{t,u,v}$, the LFTs of two consecutive frames $x_{t-1}$ and $x_t$, the *local phase difference* is then defined element-wise as:

$$\mathcal{PD}_{t-1,u,v} := \frac{\mathcal{X}_{t,u,v}\overline{\mathcal{X}_{t-1,u,v}}}{|\mathcal{X}_{t,u,v}\overline{\mathcal{X}_{t-1,u,v}}|}. \quad (7)$$

The terms above are understood to be implicitly indexed by $[\omega_1,\omega_2]$, which is omitted for convenience of notation. The inverse FFT of $\mathcal{PD}_{t-1,u,v}$ yields a corresponding cross-correlation matrix $\text{PD}_{t-1,u,v}[k,l]$, and when a perfect circular shift is observed locally, this turns out to be a

**Algorithm 1:** Local Fourier Transform - LFT
**Data:** batch of images $x$, shaped $B \times U \times V$, window function $w$, shaped $N \times N$
**Parameters:** hop size H, padding size P
**Result:** batch of *LFT* results $\mathcal{X}_{u,v}$, shaped $B \times L \times N' \times N' \times 2$
$x_{u,v} \leftarrow$ extract_local_windows($x$, $N$, $H$)
$x_{u,v} \leftarrow x_{u,v} \cdot w$
$x_{u,v} \leftarrow$ zero_pad($x_{u,v}$, $P$)
$\mathcal{X}_{u,v} \leftarrow$ FFT($x_{u,v}$)
**return** $\mathcal{X}_{u,v}$

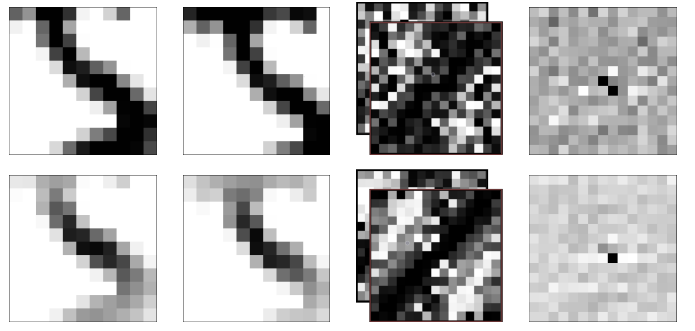

Fig. 2: *Upper Row, from left to right*: a local view on frame $x_0$, the view at the same spatial index on $x_1$, the noisy phase difference extracted between them, and the resulting ambiguous cross-correlation.
*Lower Row*: the same data when cells are tapered by a confined Gaussian window, the resulting cross-correlation now encoding a distinct shift towards the lower right.

Kronecker delta function $\delta_{t-1,u,v}[k+\Delta k, l+\Delta l]$ encoding the linear movement within a cell. Analysis of circular shifts by this method is referred to as *Phase-Only Correlation* [16].

While the description of image transforms as locally linear shifts is not restrictive, the assumption of circular cell boundaries, or in the same vein, presupposing a periodic input signal to the FFT, is. We cannot safely relax these assumptions, as resulting noise in $\mathcal{PD}_{t-1,u,v}$ is expressed as ambiguity in $\text{PD}_{t-1,u,v}[k,l]$, showing that the local translation is no longer encoded. To attenuate this shortcoming, especially for small window sizes, it is necessary to smoothly taper the intensity values towards the edge of each cell. This is demonstrated in Fig. 2. For the STFT, many appropriate window functions exist in the literature, which we can utilize in 2D via their outer products, thus conserving their benefits due to this approach's separability. Specifically, we set $w$ to the outer product version of a Gaussian or *confined Gaussian Window* as introduced in [17].

If the transformation from $x_t$ to $x_{t+1}$ were spatio-temporally constant, that is to say $\widehat{\mathcal{PD}}_{t,u,v} = \mathcal{PD}_{t-1,u,v}$, applying Fourier shift theorem yields that the Fourier partners of the local views on $x_{t+1}$, can be calculated by the element-wise product:

$$\widehat{\mathcal{X}}_{t+1,u,v} = \mathcal{X}_{t,u,v} \cdot \widehat{\mathcal{PD}}_{t,u,v}. \quad (8)$$

Ideally, this amounts to adding the phase differences in $\widehat{\mathcal{PD}}_{t,u,v}$ to the phases in each frequency bin in $\mathcal{X}_{t,u,v}$. Therefore, we refer to this operation as *local phase addition*. In practice, zero-padding of $x_{t,u,v}$ with a *padding size P* prevents wrap-around effects that can occur at this stage, as proposed by [18]. From now on and for this reason, we assume that cells feature a side length of $N' := N + 2P$.

Utilizing the inverse FFT, cells covering $x_{t+1}$ are recovered as:

$$C = e^{j\frac{2\pi}{N'}(\omega_1(n-uH)+\omega_2(m-vH))} \quad (9)$$

$$\hat{x}_{t+1,u,v}[n,m] = \frac{1}{N'^2} \sum_{\omega_1=0}^{N'-1} \sum_{\omega_2=0}^{N'-1} \widehat{\mathcal{X}}_{t+1,u,v}[\omega_1,\omega_2] C \quad (10)$$

Calculating $\hat{x}_{t+1,u,v}$ yields local predictions, but they are modified by the extraction window $w$, rendering the recovery of the global prediction $x_{t+1}$ non-trivial. Perfect reconstruction of $x_{t+1}$ given $\hat{x}_{t+1,u,v}$ and $w$ is therefore an important consideration. Keeping in mind that $\hat{x}_{t+1,u,v}$ is the result of the inverse FFT of $\widehat{\mathcal{X}}_{t+1,u,v}$, the process of perfect reconstruction is equivalent to inverting the LFT.

In analogy to the 1D case [19], one can derive and regard to this end:

$$\sum_{u=-\infty}^{\infty} \sum_{v=-\infty}^{\infty} \hat{x}_{t+1,u,v}[n,m] w^a[n-uH, m-vH]$$
$$= x_{t+1}[n,m] \sum_{u=-\infty}^{\infty} \sum_{v=-\infty}^{\infty} w^{a+1}[n-uH, m-vH].$$

Rearranging the first and the last terms yields:

$$x_{t+1}[n,m] = \frac{\sum\limits_{u=-\infty}^{\infty} \sum\limits_{v=-\infty}^{\infty} \hat{x}_{t+1,u,v}[n,m] w^a[n-uH, m-vH]}{\sum\limits_{u=-\infty}^{\infty} \sum\limits_{v=-\infty}^{\infty} w^{a+1}[n-uH, m-vH]}$$
(11)

Eq. 11 is the *overlap-add equation* for the LFT at the core of the inverse Local Fourier Transform (iLFT). In our implementation, we set $a = 1$ to avoid interpreting potential terms including $0^0$ in the numerator of equation (11). The formulation above assumes that $P$ pixels of padding around $\hat{x}_{t+1,u,v}$ have been pruned. When $P$ is at least half as large as the maximum pixel velocity present in the scene, we expect these pruned pixels to absorb and discard local wrap-around effects.

However, this stage offers an opportunity for improvement. The main dilemma in designing these systems is posed in the selection of an appropriate window size. When $N$ is too large, multiple distinct translations are potentially observed within a cell, dulling the phase differences. When it is smaller, this scenario is less likely, but the overlap between cells decreases. This is problematic in its own way, as the regions of overlap are our primary tool for transporting dynamic image content between cells. One might consider countering this by using a dense stride. However, this increases the number of extracted cells quadratically and would not be necessary if not for the benefit of increasing cell overlap. We suggest a compromise: instead of using smaller strides, increase $P$ slightly to match the maximum scene pixel velocity. Before the reconstruction step, we do not prune this padding but instead create locally adaptive versions of the window function by zero-padding $w$ and calculating its *FFT* $\mathcal{W}$ and finally, modify it locally via:

$$\hat{w}_{t,u,v} := \text{iFFT}(\text{phase\_add}(\mathcal{W}, \widehat{\mathcal{PD}}_{t,u,v})), \tag{12}$$

which corresponds to shifting it according to the locally observed transformations. This can be summarized by replacing $w$ with $\hat{w}_{t,u,v}$ in equation 11, thereby implementing the *modified inverse Local Fourier Transform* described in algorithm 2.

---
**Algorithm 2:** modified inverse Local Fourier Transform - m_iLFT

**Data:** batch of LFT results $\widehat{\mathcal{X}}_{u,v}$, shaped
$B \times L \times N' \times N' \times 2$,
window function $w$, shaped $N \times N$
batch of phase differences $\widehat{\mathcal{PD}}_{t,u,v}$, shaped
$B \times L \times N' \times N' \times 2$
**Parameters:** hop size H, padding size P
**Result:** batch of images $x$, shaped $B \times U \times V$
$\hat{x}_{u,v} \leftarrow \text{iFFT}(\widehat{\mathcal{X}}_{u,v})$
$w \leftarrow \text{zero\_pad}(w, P)$
$\mathcal{W} \leftarrow \text{FFT}(w)$
$\widehat{\mathcal{W}}_{t,u,v} \leftarrow \text{phase\_add}(\mathcal{W}, \widehat{\mathcal{PD}}_{t,u,v})$
$\hat{w}_{t,u,v} \leftarrow \text{iFFT}(\widehat{\mathcal{W}}_{t,u,v})$
$\hat{x}_{u,v} \leftarrow \hat{x}_{u,v} \cdot \hat{w}_{t,u,v}$
$num \leftarrow \text{overlap\_add}(\hat{x}_{u,v}, H)$
$denom \leftarrow \text{overlap\_add}(\hat{w}^2_{t,u,v}, H)$
$x \leftarrow \frac{num}{denom}$
**return** $x$

---

Effectively, we have obtained windows within cells that stay focused on moving image contents between time steps, ensuring they are represented in reconstruction even when they move into the field of view of a nearby cell. The main disadvantage of this modification to image synthesis is that the denominator in equation 11 is no longer inherently safe. Of course, this term must be nonzero within the support of $x_{t+1}$, which is sometimes referred to as the *nonzero overlap-add* or *NOLA condition* in the context of the STFT [20]. This is a property of the window function together with the stride and is readily anticipated in the unmodified term. With shifted windows, however, it can only be asserted at runtime. In practice, this does not cause any problems unless the window is pathological in the sense that it has minuscule support.

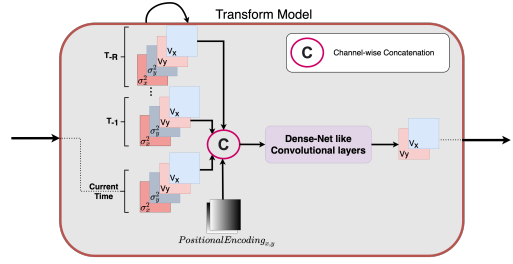

Fig. 3: The "transform model" filters local phase differences using DenseNet-like convolutional layers.

In practice, we also need to revisit $\widehat{\mathcal{PD}}_{t,u,v} = \mathcal{PD}_{t-1,u,v}$ before Eq. 8, the assumption that the transformation between frames $x_{t-1}$ and $x_t$ is applicable to predict the motion towards the consecutive frames. This is exclusive to farcical scenarios, such as objects carried in a homogeneous stream. In most scenes, objects display their own motility, and the corresponding local shifts must be carried along with them. Otherwise, predictions are doomed to break down because each of the extracted cells' receptive fields is limited and cannot grasp the full movement pattern. Fortunately, this spatiotemporal transport of the transformation is not arbitrary for short time intervals and occurs systematically following the object movement. We use a small convolutional model to have a bigger receptive field than a single small local window, and by aggregating them, the model can reason about more global movement. This model has a bigger receptive field than a single local window; hence it is sufficient to approximate and understand a coherent movement. To denote this, we write $\widehat{\mathcal{PD}}_{t,u,v} = \mathcal{MM}(\mathcal{PD}_{t-1,u,v})$, where $\mathcal{MM}$ is called the "transform model". Including it between the previous steps completes the prediction process as outlined in algorithm 3. This process is also illustrated in Fig. 1, where analysis of phase differences in the context of a prediction step is visualized for two arbitrary cells accentuated by red and blue borders.

---
**Algorithm 3:** predict_next_frame

**Data:** batch of images $x_{t-1}$, shaped $B \times U \times V$
batch of images $x_t$, shaped $B \times U \times V$,
window function $w$, shaped $N \times N$
**Parameters:** hop size H, padding size pS
**Result:** batch of images $x_{t+1}$, shaped $B \times U \times V$
$\mathcal{X}_{t-1,u,v} \leftarrow \text{LFT}(x_{t-1}, w)$
$\mathcal{X}_{t,u,v} \leftarrow \text{LFT}(x_t, w)$
$\mathcal{PD}_{t-1,u,v} \leftarrow \text{get\_phase\_differences}(\mathcal{X}_{t,u,v}, \mathcal{X}_{t-1,u,v})$
$\widehat{\mathcal{PD}}_{t,u,v} \leftarrow \mathcal{MM}(\mathcal{PD}_{t-1,u,v})$
$\widehat{\mathcal{X}}_{t+1,u,v} \leftarrow \text{phase\_add}(\mathcal{X}_{t,u,v}, \widehat{\mathcal{PD}}_{t,u,v})$
$x_{t+1} \leftarrow \text{m\_iLFT}(\widehat{\mathcal{X}}_{t+1,u,v}, w, \widehat{\mathcal{PD}}_{t,u,v})$
**return** $x_{t+1}$

---

Without further consideration, manipulating individual frequency bins via unregularized and unstructured learn-ing models exhibits an obscure nature. In a freely evolving learning model, and if we want to merely optimize $\widehat{\mathcal{PD}}_{t,u,v}$ for the prediction loss, image contents can be altered in unanticipated ways when Eq. 8 is applied in the frequency domain. To guard against this forfeiture of interpretability and prevent the model from optimizing purely in an opportunistic manner, we ensure that the output of $\mathcal{MM}$ encodes exclusively local shift and does not alter other aspects of the prediction by introducing a bottleneck to the system.

### A. Transform model

If we assume that each of the local phase differences only encodes one local shift, we can create a bottleneck in the pipeline, in which each of the local phase differences is described by two numbers. We call it local velocity $V$, and visualize it using little red arrows like in, for example, Fig. 6. To extract $V$ from local phase differences, we compute the average of the adjacent elements' differences in the $x$ and $y$ direction. This is a weighted average operation accounting for the energy of the underlying frequency bin. The result is two complex numbers $M_x$ and $M_y$ representing velocity in each of the two directions. We can convert these two complex numbers to the local pixel velocities:

$$V = \begin{bmatrix} V_x \\ V_y \end{bmatrix} = \begin{bmatrix} \frac{N \times atan2(\Re(M_x), \Im(M_x))}{2\pi} \\ \frac{N \times atan2(\Re(M_y), \Im(M_y))}{2\pi} \end{bmatrix} \quad (13)$$

Next, we reshape these local velocities to the shape $L_U \times L_V \times 2$. In the "transform model", we use a few learnable parameters in the form of DenseNet-like convolutional layers to filter these local velocities.

The "transform model" filters local velocities by accounting for the neighbor's local velocities. To get a better result and help the model decide how reliable each local velocity is, we also include each direction's variance $[\sigma_x^2, \sigma_y^2]$ and concatenate it channel-wise. Since a convolutional model cannot learn location-dependent features, similar to the positional encoding proposed by Azizi et.al [21], we add two additional channels to the input. To grant the model the ability account for former velocities, we channel-wise concatenate the same saved features from previous time steps. If we consider $R$ previous time steps, the transform model's input shape is $L_U \times L_V \times (2+2)*(R+1)+2$.

The network's output is two refined velocities $\widehat{V}$ with the shape of $L_U \times L_V \times 2$. Since the network's output is in pixel speed, we now have to convert $\widehat{V}$ back and re-create local phase differences. To do so, we first convert each element of $\widehat{V}$ to $\widehat{M}$, which is an angle in the polar representation of the complex number:

$$\widehat{M} = \begin{bmatrix} \widehat{M_x} \\ \widehat{M_y} \end{bmatrix} = \begin{bmatrix} \frac{2\pi \widehat{V_x}}{N} \\ \frac{2\pi \widehat{V_y}}{N} \end{bmatrix} \quad (14)$$

To streamline the creation of local phase differences, we compute a template matrix $\mathcal{TM}$ once. To create $\mathcal{TM}$ we start with an empty matrix shaped $N \times N \times 2$. To fill the complex matrix $\mathcal{TM}$, we iterate over all elements, starting from the central element $Mid$, which corresponds to the zero-frequency bin. $Mid$ is set to $1+0i$ and then starting from $Mid$, we iterate through $\mathcal{TM}$ and fill each element according to:

$$\mathcal{TM}_{i,j\pm 1} = \begin{bmatrix} \Re(\mathcal{TM}_{i,j}) \\ \Im(\mathcal{TM}_{i,j}) \pm 1 \end{bmatrix}, \mathcal{TM}_{i\pm 1,j} = \begin{bmatrix} \Re(\mathcal{TM}_{i,j}) \pm 1 \\ \Im(\mathcal{TM}_{i,j}) \end{bmatrix} \quad (15)$$

The result is a matrix that is constructed once and cached for use at every window location in each iteration. The final local phase difference result is:

$$\widehat{\mathcal{PD}} = \begin{bmatrix} \Re(\widehat{\mathcal{PD}}) \\ \Im(\widehat{\mathcal{PD}}) \end{bmatrix} = \begin{bmatrix} \cos(\widehat{M_x} \Re(\mathcal{TM})) \\ \sin(\widehat{M_y} \Im(\mathcal{TM})) \end{bmatrix} \quad (16)$$

To summarize, as shown in Fig. 3, the input to the "transform model" is local phase differences $\mathcal{PD}$, and the output is filtered local phase differences $\widehat{\mathcal{PD}}$.

## IV. Motion Segmentation

We utilize LFDTN as a building block to extend the motion segmentation architecture recently proposed by Farazi et al. [12]. This extension enables the motion segmentation architecture to model non-global transformations and demonstrates the efficacy of our proposed LFDTN as a building block in more advanced video prediction pipelines.

This motion segmentation pipeline models foreground and background separately while necessitating minimum trainable parameters, and it is fully interpretable.

Our motion segmentation pipeline is inspired by Kalman filters. In a Kalman filter, $x_t$ is a noisy linear function of the previous time step state $x_{t-1}$. The observation $z_t$ is modeled as a noisy linear function of the state $x_t$:

$$\begin{aligned} \mathbf{x}_t &= \mathbf{F}\mathbf{x}_{t-1} + \text{Noise} \\ \mathbf{z}_t &= \mathbf{H}\mathbf{x}_t + \text{Noise} \end{aligned}, \quad (17)$$

where $\mathbf{H}$ is the measurement matrix, and $\mathbf{F}$ is the state-transition matrix. Using these presuppositions, the posterior estimate of the state $x_t$ is calculated by:

$$\widehat{\mathbf{x}}_t = \underbrace{\widehat{\mathbf{x}}_{t|t-1}}_{\text{prediction}} + \underbrace{\mathbf{K}(\mathbf{z}_t - \widehat{\mathbf{z}}_{t|t-1})}_{\text{correction}}, \quad (18)$$

where $\mathbf{K}$ is the Kalman gain matrix, which controls how much we rely on the current prediction $\widehat{x}_{t|t-1}$ versus the observation $z_t$. Also, given observations $z_1, \ldots, z_{t-1}$; $\widehat{x}_{t|t-1}$ and $\hat{z}_{t|t-1}$ are the predictions of $x_t$ and $z_t$, respectively.

Fig. 4 demonstrates our model for self-supervised motion segmentation. We represent foreground ($\mathsf{FG_t}$) and background ($\mathsf{BG_t}$) separately as images having the same shape as the observed frames ($\mathsf{F_t}$). We then combine them

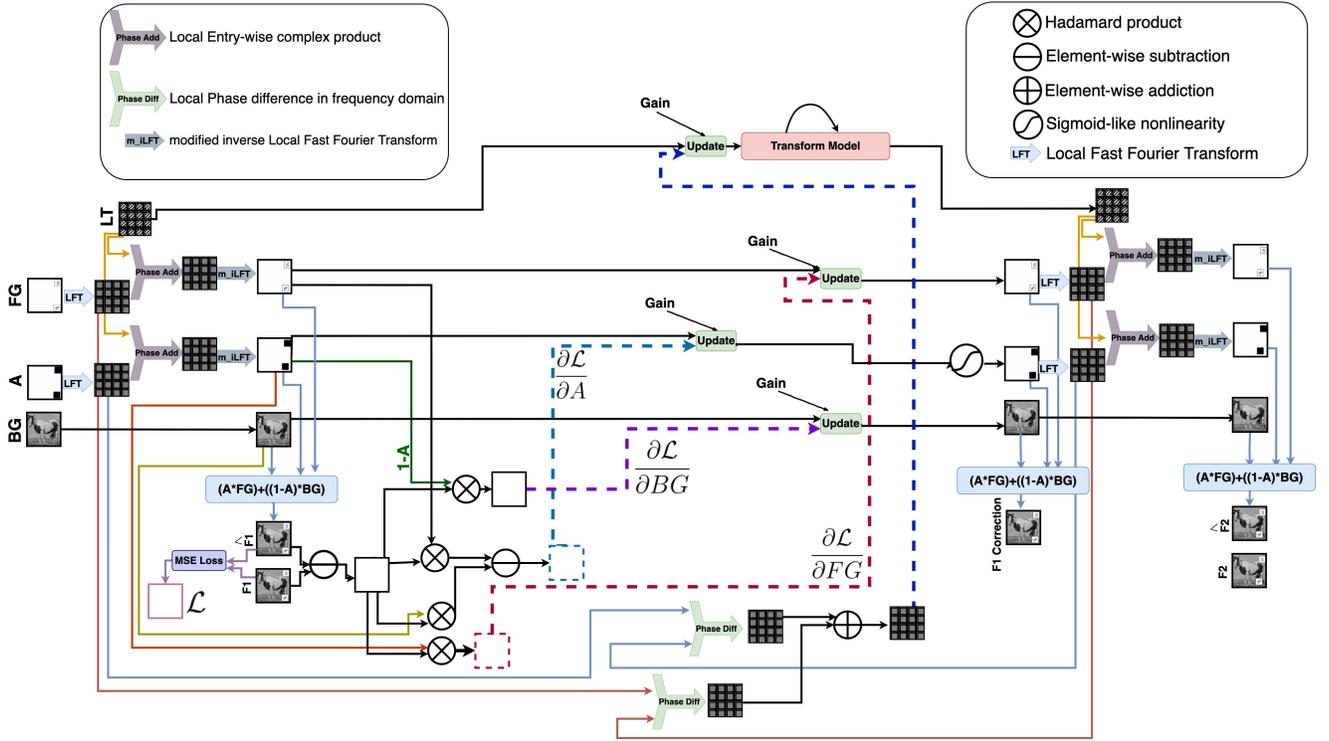

Fig. 4: Our proposed motion segmentation model. Foreground (FG) and background (BG) are modeled separately and combined using an alpha mask (A) to form the predicted next frame $\widehat{F1}$, which is later compared to the input frame F1. The prediction error is used to update FG, BG, and A. For FG and A, motion is estimated by computing local phase differences (LT). This motion estimate is added to the local phases of FG and A to locally move them accordingly. After a few seed steps of this prediction-correction cycle, the model does not need the input frames anymore and can continue predicting using only the estimated state (FG, BG, A, LT).

to model occlusions of the background by the foregrounds using the alpha mask $\widehat{A}_t$:

$$\widehat{F}_t = \widehat{A}_t \cdot \widehat{FG}_t + (1 - \widehat{A}_t) \cdot \widehat{BG}_t. \tag{19}$$

In addition to these three image-like states, the state also consists of the estimated joint local movement speeds $LT_t$ of foreground and alpha mask.

As described in the previous section, $LT_t$ is represented as local phase differences between consecutive frames in the Fourier domain. Like in Sec. III, the next foreground frame $(\widehat{FG}_t, \widehat{A}_t)$ can be predicted by phase-adding refined $LT_t$ to the local Fourier representations $LFT(.)$ of $(FG_{t-1}, A_{t-1})$. After going back to the spatial domain by the modified inverse Local Fourier transformation $m\_iLFT^{-1}(.)$, the foreground and alpha mask are moved according to the estimated local movement velocities.

We compute the difference between the predicted frame $\widehat{F}_t$ and perceived frame $F_t$ and update each part of the state by minimizing the mean squared loss $\mathcal{L}(\widehat{F}_t, F_t)$. As a simple differentiable function graph computes the predicted frame, we can efficiently perform gradient descent using a function graph for the backward pass with the same structure. Instead of updating each state using automatic differentiation packages, we hard-wired gradient computation in the computational graph. This results in a computation graph that obtains a Kalman filter-like prediction-correction cycle in its forward pass. For updating the state $LT_t$, which is in local Fourier space, we calculate the phase differences between $LFT(A_t)$ and $LFT(A_{t-1})$ as well as $LFT(FG_t)$ and $LFT(FG_{t-1})$. For updating $LT_t$, we take a weighted average between $LT_{t-1}$ and the estimated local phase differences $\widehat{LT}_t$. To train faster and prevent the network from grouping everything into the FG in the early stages of training, we add an $L1$ regularization of the A to the loss term.

The introduced model pre-structures our assumptions that foreground objects move in front of a stationary background and occlude the background according to the alpha mask. Furthermore, we also hard-wire motion estimation and prediction by using LFDTN (Sec. III).

So far, our segmentation design does not append any learnable parameters to LFDTN, so we cannot learn to exploit the statistical characteristics of data efficiently. We can backpropagate a loss through the unfolded network in time because our prediction-correction computation graph is fully differentiable.

Hereafter, any parameter can be updated by gradient descent, and we can easily add parameters at appropriate computation steps. For initializing the spatial states FG, BG, and A we use three separate convolutional networks.

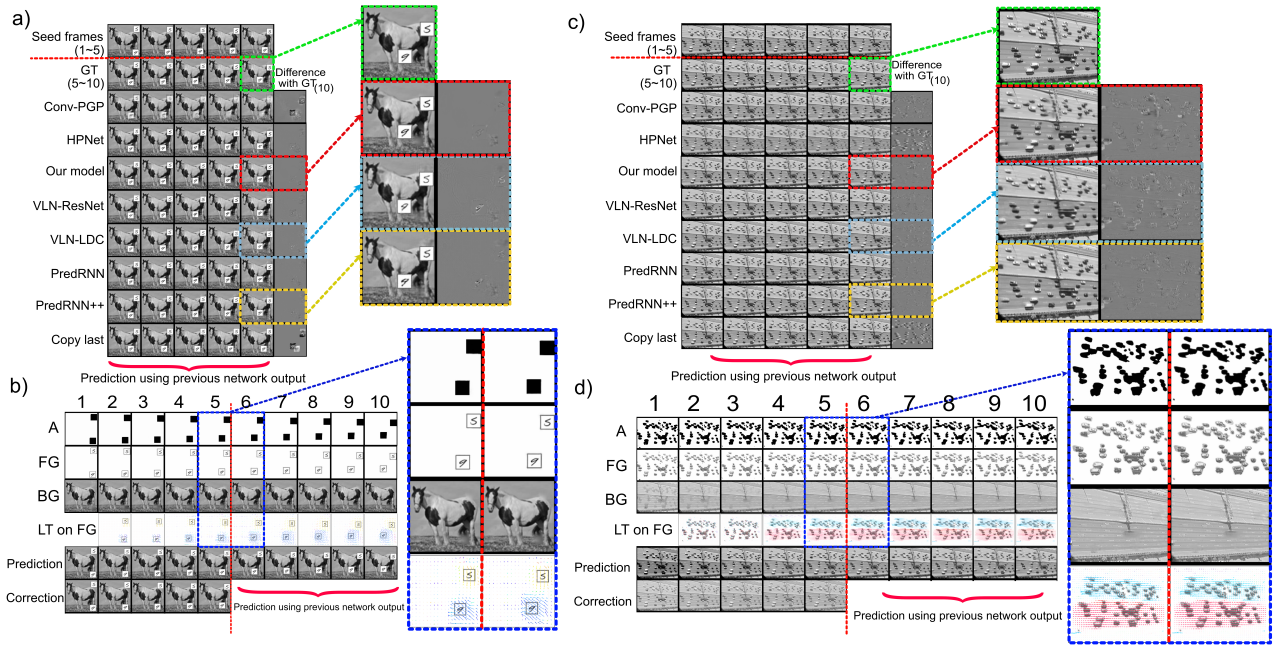

Fig. 5: a) Predictions for a randomly selected sample with different models on "Moving MNIST on STL". b) Internal states' development in our model. Note that our model can segment foreground and background and estimate foreground motion. c) Predictions for a randomly selected sample with different models on the "NGSIM". d) Internal states' development in the motion segmentation model. Note due to padding around the image; the LT are depicted on a larger canvas than other states. The animated and full-resolution version of all sample results can be found online[†].

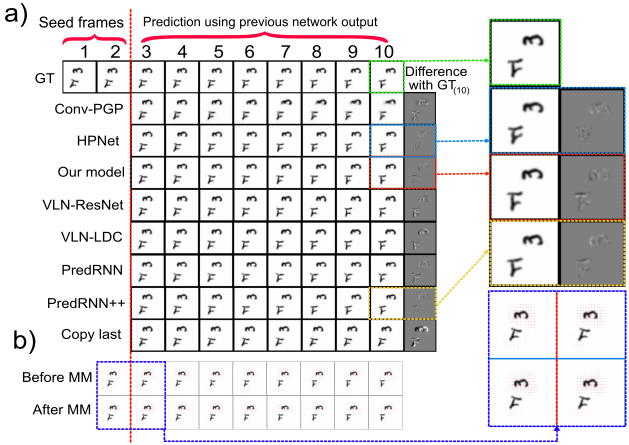

Fig. 6: a) Predictions for a randomly selected sample with different models on the "Moving MNIST++" dataset. b) Internal local movement representation in our LFDTN, before and after "Transform Model". Due to padding, the frame for internal representation of LT is larger than pure prediction output. More can be seen online[†].

Depending on the dataset's difficulty, each has $D$ convolutional layers, with DenseNet-like connections, followed by PReLU activations. $D$ can range from three in the simple datasets to eight in more difficult datasets. We also initialize LT by combining the $LFT(.)$ of two initial steps of A and FG. Each state is the weighted average among the updated state and the convolutional network's output in the first couple of steps. We use a decaying gain for this weighted average such that in the initial step, we only use the convolutional network output, and later we rely more and more on the updated states. Note that the convolutional network also fills-in occluded parts of the background BG.

## V. Experimental Results

Our proposed models and the update gains in the "motion segmentation" model are trained end-to-end using backpropagation through time. We used AdamW optimizer [22] and a hybrid combination of SSIM and discounted MSE prediction loss supplemented by a cyclic learning rate schedule.

To evaluate our models, we used three different datasets, two of which are synthetic.

We use a variant of the Moving MNIST data set to evaluate our proposed architecture, which we call "Moving MNIST++". Each sample contains ten frames with two MNIST images, moving inside a 64×64 frame. In addition to translation in classical "Moving MNIST", it also contains more difficult transformations like rotation and scaling. In each experiment, the first two frames were seed frames, and the rest was predicted. We evaluated our LFDTN model on this dataset. We compared our model against many well-known models like Conv-PGP [8], VLN-ResNet [23], VLN-LDC [21], HPNetT [6], PredRNN [24],

[†]http://ais.uni-bonn.de/~hfarazi/LFDTN/

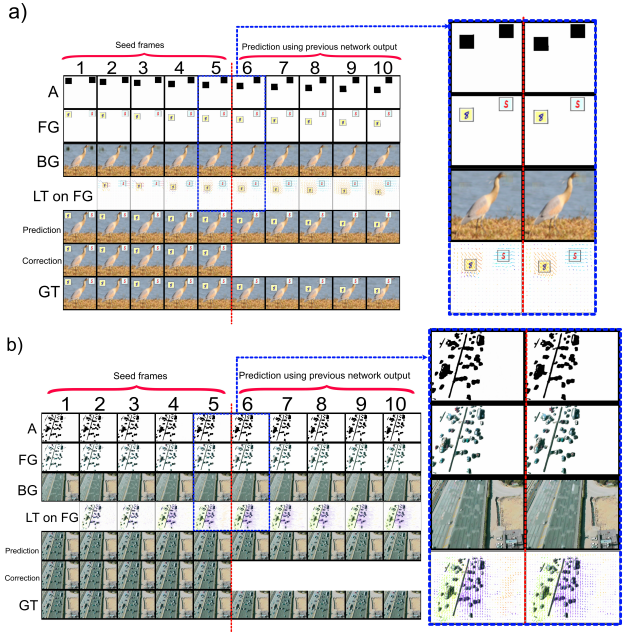

Fig. 7: Internal states' formation for two samples for color a) "Moving MNIST on STL", and b) "NGSIM" datasets. Note that the depicted frames for LT, are more widespread than other states due to padding. See more online[†].

and PredRNN++ [4]. We also showed the result if we simply copy the last seed frame.

In this experiment, the only used learnable parameters were two convolutional layers of size 3 in the "transform model". Sample results of our models, as well as baselines, are presented in Fig. 6. Since we can visualize the estimated local movements in the form of local velocity arrows, the formed representations are easily interpretable in Fig. 6. Table I reports the prediction losses, structural (dis-)similarity, and the number of parameters for the evaluated models. It can be observed that our proposed model outperforms our baselines in almost all metrics with far less parameters.

We argue that MSE is not a useful metric for video prediction since the model can essentially cheat by blurring the output to receive a good MSE score. Since both of our models by design are unlikely to produce blurry results, it is expected that models with blurry predictions can potentially outperform our models on this metric. Our models are less likely to make blurry predictions because both models can only locally shift moving patches of objects. Moreover, the motion segmentation model has to draw a sharp border of foreground objects using the alpha mask. We also think BCE can be a useful metric for training and testing binary signals, but it is not suitable to represent a non-binary dataset. Since it is common to show BCE and MSE metrics in other video prediction works, we covered them here. We believe that SSIM is the best metric for evaluating video prediction results.

The second synthetic dataset, which we call "Moving MNIST on STL", has two randomly selected MNIST digits moving with subpixel accuracy on a randomly selected STL-10 image. Sample results of our models, as well as other baselines, are displayed in Fig. 5. Table II reports the outcomes. It can be seen that our proposed model defeats our baselines in almost all metrics with a notably lower number of parameters. The internal states formation of our model is represented in Fig. 5.

The third dataset was obtained using raw traffic camera footage from Interstate 80 Freeway Dataset of Next Generation Simulation (NGSIM) 5. Sample results of our models, as well as other used baselines, can be seen in Fig. 5. Table III summarizes the outcomes. Our proposed model outperforms other models on SSIM and L1 metrics. The formation of internal states of our model is shown in Fig. 5. It is worth noting that our model uses orders of magnitude fewer learnable parameters. In this experiment, we utilized progressive, growing training to train our model faster and more efficiently.

In the last two experiments, we used four DenseNet-like convolutional layers in the "transform model" and eight layers to initialize the models' spatial states.

In stark contrast to other baselines, for both the "Moving MNIST on STL" and "NGSIM" datasets, we can visualize and understand the internal representation; the model can also segment foreground and background without any supervision, solely by minimizing the self-supervised prediction loss. This is a direct consequence of cementing the occlusion assumptions and Kalman filter inspired update cycle in our model.

Overall all three experiments point to the fact that our proposed model can perform very well compared to other video prediction methods while having a fraction of learnable parameters and without sacrificing interpretability.

We use grayscale images on all three datasets to evaluate our models against available baseline codes without massively extending and changing them. To show that with minimal changes, our models can predict on RGB images, we created a colored version of "Moving MNIST on STL" and RGB color full resolution "NGSIM" from raw footage. The only necessary change to our model in order to work on the colored dataset was that we estimate joint movements on all channels and apply the LFDTN result per channel. For motion segmentation, the state, FG, and BG is also stored with three channels. You can see sample results as well as formed internal representation in Fig. 7.

VI. Conclusion and Future Work

We proposed Local Frequency Transformer Networks, a fully interpretable and lightweight differentiable module for the video prediction task. This network can estimate local velocities, project them into the future, and transform content using projected velocities. We also proposed an end-to-end learnable network architecture for motion segmentation and video prediction using LFDTN. This network estimates interpretable internal states using a prediction-correction scheme. It needs very few learnable

TABLE I: Prediction losses for "Moving MNIST++"*.

| Model | L1 | MSE | DSSIM | BCE | # of Params |
|---|---|---|---|---|---|
| Conv-PGP [8] | 0.02066 | 0.00357 | 0.09981 | 0.07500 | **32K** |
| HPNET [6] | 0.00681 | 0.00075 | 0.00715 | 0.07113 | 15.8M |
| Our LFDTN | **<u>0.00589</u>** | 0.00067 | **<u>0.00656</u>** | **<u>0.06299</u>** | **<u>3K</u>** |
| VLN-ResNet [23] | 0.01330 | 0.00308 | 0.02185 | 0.06840 | 1.3M |
| VLN-LDC [21] | 0.01293 | 0.00289 | 0.02052 | 0.06785 | 1.3M |
| PredRNN [24] | 0.00892 | 0.00122 | 0.01259 | 0.07226 | 1.8M |
| PredRNN++ [4] | **0.00616** | **<u>0.00064</u>** | **0.00660** | **0.06392** | 2.8M |
| Copy last frame | 0.03270 | 0.01528 | 0.07067 | 0.27600 | - |

TABLE II: Prediction losses for "Moving MNIST on STL"*.   TABLE III: Prediction losses for "NGSIM"*.

| Model | L1 | MSE | DSSIM | BCE | Params | L1 | MSE | DSSIM | BCE | # of Params |
|---|---|---|---|---|---|---|---|---|---|---|
| Conv-PGP [8] | 0.0347 | 0.0072 | 0.06464 | 0.5385 | **<u>32K</u>**. | 0.0323 | 0.0037 | 0.0713 | 0.6029 | **153K** |
| HPNET [6] | 0.0097 | 0.0013 | 0.01613 | 0.5483 | 15.8M | 0.0571 | 0.0110 | 0.1224 | 0.6196 | 15.8M |
| Our Motion Seg | **<u>0.0049</u>** | **<u>0.0005</u>** | **<u>0.0067</u>** | 0.5174 | 82K | **<u>0.0229</u>** | 0.0027 | **<u>0.0457</u>** | 0.6018 | **<u>91K</u>** |
| VLN-ResNet [23] | 0.01626 | 0.0012 | 0.0268 | 0.5178 | 1.3M | 0.0330 | 0.0032 | 0.0835 | 0.6017 | 1.3M |
| VLN-LDC [21] | 0.01419 | 0.0009 | 0.0189 | **<u>0.5171</u>** | 1.4M | 0.0323 | 0.0032 | 0.0830 | **<u>0.6017</u>** | 1.5M |
| PredRNN [24] | 0.0101 | 0.0006 | 0.0142 | 0.5224 | 4M | 0.0277 | **0.0023** | 0.0654 | 0.6021 | 14.9M |
| PredRNN++ [4] | **0.0091** | **0.0006** | **0.0131** | 0.5214 | 6.3M | **0.0241** | **<u>0.0018</u>** | **0.0521** | **0.6017** | 23.2M |
| Copy last frame | 0.0315 | 0.0190 | 0.0504 | 1.7869 | - | 0.0358 | 0.0099 | 0.0819 | 0.6186 | - |

*: Note on all above tables the best result on each metric is marked by making the numbers **<u>bold and underline</u>**, while the second-best is marked by making it **bold**.

parameters, making it sample efficient and highly generalizable to unforeseen data. Experiments on synthetic and real data indicate that with far fewer parameters, both of our methods can perform very well compared to other baselines.

For the future, we want to extend our LFDTN to utilize multiple levels of granularity. We also plan to improve our motion segmentation model to have more depth channels, accounting for those instances in which foreground objects also occlude each other.

### Acknowledgment

This work was funded by grant BE 2556/16-1 (Research Unit FOR 2535 Anticipating Human Behavior) of the German Research Foundation (DFG). The authors would like to thank the open-source community and Mark Prediger for providing code for some of the baseline methods. The authors would also like to thank W&B [25] for providing free HPO and monitoring tools.

### References

[1] H. Farazi and S. Behnke, "Frequency domain transformer networks for video prediction," in *ESANN*, 2019.
[2] F. Cricri, X. Ni, M. Honkala, E. Aksu, and M. Gabbouj, "Video ladder networks," *CoRR*, vol. abs/1612.01756, 2016.
[3] A. Rasmus, M. Berglund, H. Valpola, and T. Raiko, "Semi-supervised learning with ladder networks," in *NIPS*, 2015.
[4] Y. Wang, Z. Gao, M. Long, J. Wang, and P. S. Yu, "Predrnn++: Towards a resolution of the deep-in-time dilemma in spatiotemporal predictive learning," *ArXiv*, vol. abs/1804.06300, 2018.
[5] W. Lotter, G. Kreiman, and D. D. Cox, "Deep predictive coding networks for video prediction and unsupervised learning," *CoRR*, vol. abs/1605.08104, 2016.
[6] J. Qiu, G. Huang, and T. Lee, "A neurally-inspired hierarchical prediction network for spatiotemporal sequence learning and prediction," *ArXiv*, vol. abs/1901.09002, 2019.
[7] V. Michalski, R. Memisevic, and K. Konda, "Modeling deep temporal dependencies with recurrent grammar cells," in *NIPS*, 2014.
[8] F. D. Roos., "Modeling spatiotemporal information with convolutional gated networks," Master's thesis, Chalmers University of Technology, 2016.
[9] R. Memisevic, "Learning to relate images: Mapping units, complex cells and simultaneous eigenspaces," *ArXiv*, vol. abs/1110.0107, 2011.
[10] J. R. van Amersfoort, A. Kannan, M. Ranzato, A. Szlam, D. Tran, and S. Chintala, "Transformation-based models of video sequences," *CoRR*, vol. abs/1701.08435, 2017.
[11] E. Ilg, N. Mayer, T. Saikia, M. Keuper, A. Dosovitskiy, and T. Brox, "Flownet 2.0: Evolution of optical flow estimation with deep networks," *CVPR*, pp. 1647–1655, 2017.
[12] H. Farazi and S. Behnke, "Motion segmentation using frequency domain transformer networks," in *ESANN*, 2020.
[13] R. Memisevic, "Learning to relate images," *IEEE transactions on pattern analysis and machine intelligence*, 2013.
[14] R. N. Bracewell, *The fourier transform and its applications /.* McGraw-Hill,, 3rd ed. ed., 2000.
[15] A. A. Lazar, N. H. Ukani, and Y. Zhou, "A motion detection algorithm using local phase information," *Computational intelligence and neuroscience*, 2016.
[16] K. Takita, T. AOKI, Y. SASAKI, T. HIGUCHI, and K. KOBAYASHI, "High-accuracy subpixel image registration based on phase-only correlation," *IEICE Transactions on Fundamentals of Electronics, Communications and CS*, 2003.
[17] S. Starosielec and D. Hägele, "Discrete-time windows with minimal rms bandwidth for given rms temporal width," *Signal Processing*, 2014.
[18] J. Allen, "Short term spectral analysis, synthesis, and modification by discrete fourier transform," *IEEE Transactions on Acoustics, Speech, and Signal Processing*, 1977.
[19] X. Li, "A simplified normalization operation for perfect reconstruction from a modified stft," in *2014 12th International Conference on Signal Processing (ICSP)*, 2014.
[20] D. Griffin and J. Lim, "Signal estimation from modified short-time fourier transform," *IEEE Transactions on acoustics, speech, and signal processing*, 1984.
[21] Niloofar Azizi, Hafez Farazi and S. Behnke, "Location dependency in video prediction," in *International Conference on Artificial Neural Networks (ICANN)*, 2018.
[22] I. Loshchilov and F. Hutter, "Decoupled weight decay regularization," *arXiv preprint arXiv:1711.05101*, 2017.
[23] F. Cricri, X. Ni, M. Honkala, E. Aksu, and M. Gabbouj, "Video ladder networks," *arXiv:1612.01756*, 2016.
[24] Y. Wang, M. Long, J. Wang, Z. Gao, and P. S. Yu, "Predrnn: Recurrent neural networks for predictive learning using spatiotemporal lstms," in *NPIS*, 2017.
[25] L. Biewald, "Experiment tracking with weights and biases," 2020. Software available from wandb.com.